\documentclass[lettersize,journal]{IEEEtran}
\usepackage{amsmath,amsfonts}
\usepackage{algorithmic}
\usepackage{algorithm}
\usepackage{array}
\usepackage[caption=false,font=normalsize,labelfont=sf,textfont=sf]{subfig}
\usepackage{textcomp}
\usepackage{stfloats}
\usepackage{url}
\usepackage{verbatim}
\usepackage{graphicx}
\usepackage{cite}
\usepackage[colorlinks=true, linkcolor=blue, urlcolor=blue]{hyperref}
\usepackage{multirow}
\usepackage{bbding}
\usepackage{booktabs}
\usepackage{colortbl}
\usepackage[table,xcdraw]{xcolor}

\begin{document}

\title{LoHGNet: Infrared Small Target Detection through Lorentz Geometric Encoding with High-Order Relation Learning}

\author{Qianwen~Ma, Yang~Xu,  Shangwei~Deng, Xiaobo~Li~\IEEEmembership{Member,~IEEE,} and Haofeng~Hu
\thanks{This work was supported by the National Natural Science Foundation of China (No. 62475190), the Open Project Funds for the Key Laboratory of Space Photoelectric Detection and Perception (Nanjing University of Aeronautics and Astronautics), Ministry of Industry and Information Technology (No. NJ2024027-1), Tianjin Science and Technology Program (Nos. 25JCJQJC00190 and 23YFZCSN00230), and the Fundamental Research Funds for the Central Universities (No. NJ2024027). (Corresponding author: X. Li.)}
\thanks{Q. Ma, S. Deng, X. Li, and H. Hu are with the School of Marine Science and Technology, Tianjin University, Tianjin 300072, China (e-mail: kingwin@tju.edu.cn, shangweideng@tju.edu.cn, lixiaobo@tju.edu.cn, haofeng\_hu@tju.edu.cn).}
\thanks{Y. Xu is with the Jiangxi Key Laboratory of Advanced Electronic Materials and Device, Jiangxi Science and Technology Normal University, Nanchang 330028, China (e-mail: 2024010291@jxstnu.edu.cn).}
}

\markboth{IEEE XXXX}%
{Shell \MakeLowercase{\textit{et al.}}: A Sample Article Using IEEEtran.cls for IEEE Journals}


\maketitle

\begin{abstract}


Infrared small target detection (IRSTD) remains challenging due to the scarcity of useful target cues and the presence of severe background clutter. Most current methods rely on conventional feature learning and local interaction modeling, where features are represented in Euclidean space. However, such designs may still be limited in describing the subtle differences of weak targets and the contextual relations between targets and backgrounds. To address these limitations, we propose LoHGNet, an IRSTD network that integrates Lorentz geometric encoding with high-order relation learning. By introducing Lorentz manifold based feature learning, LoHGNet offers a different feature representation from conventional IRSTD methods and provides new discriminative cues for IRSTD. Specifically, a Lorentz encoding branch is constructed with the Geometric Attention Guided Lorentz Residual Convolution Module (GA-LRCM) to perform feature modeling under hyperbolic geometric constraints and enhance the hierarchical geometric representation capability of weak targets. Subsequently, the hyperbolic features are mapped into the Euclidean tangent space through logarithmic mapping, and a High-Order Relation Learning Module (HORL) is designed to model the high-order contextual dependencies between targets and backgrounds via hypergraph construction, thereby improving target discrimination in complex backgrounds. Experimental results on three datasets demonstrate that the proposed LoHGNet achieves competitive performance in both detection accuracy and adaptability to complex scenes. The code will be available at \href{https://github.com/Kingwin97/}{https://github.com/Kingwin97}.

\end{abstract}


\section{Introduction}
\IEEEPARstart{I}{nfrared} small target detection (IRSTD) is a key technique in applications such as search and rescue warning, target surveillance, and precision guidance \cite{wu2026dfinet}. Different from conventional object detection tasks, infrared small targets usually occupy only a few pixels and are generally characterized by extremely small scale, limited texture and structure information, and low signal-to-noise ratio \cite{ji2026three}. In complex scenes, these properties make targets highly susceptible to being submerged by cloud edges, bright ground regions, and sensor noise, thereby substantially increasing the difficulty of feature modeling and accurate detection.

Over the past decades, IRSTD has developed into a relatively rich research field \cite{chen2026dcganet}. Early methods mainly relied on infrared imaging mechanisms and prior knowledge tailored to specific scenarios. Although such methods can be effective under controlled conditions, manually designed features and fixed hyperparameters are difficult to adapt to diverse imaging conditions in practical scenarios with complex background structures, severe noise interference, or low target saliency. With the continuous development of the data-driven paradigm, convolutional neural network methods have gradually become the dominant paradigm in IRSTD, significantly improving feature representation and target detection performance in complex scenes. 

However, existing methods still face a more fundamental challenge in small target detection, as the separation between true targets and complex backgrounds is often insufficient. Small targets usually occupy only a few pixels, and their responses are weak. In complex scenes, they can also appear locally similar to noise or bright clutter. In this case, detection does not depend only on the response intensity itself, but also on subtle and hierarchical structural differences between the target and the background. Most existing networks learn features in Euclidean space. Since Euclidean space has a flat and uniform metric structure, it is more suitable for describing locally continuous changes, but less effective for representing the non-uniform and hierarchical geometric relations between weak targets and complex backgrounds. As a result, during deep propagation and repeated downsampling, weak target-related structural cues are easily diluted by dominant background information, making it difficult to enlarge the representation gap between true targets and false responses. Meanwhile, most existing methods model contextual relations mainly through local receptive fields or pairwise interaction mechanisms. Such designs are still limited in modeling high-order dependencies among multiple regions, which makes it harder to distinguish real targets from false responses caused by the background \cite{ciocarlan2026anomaly, wu2026semdetnet}.

\begin{figure*}[h]  
  \begin{center}
  \includegraphics[width= 1.0\linewidth]{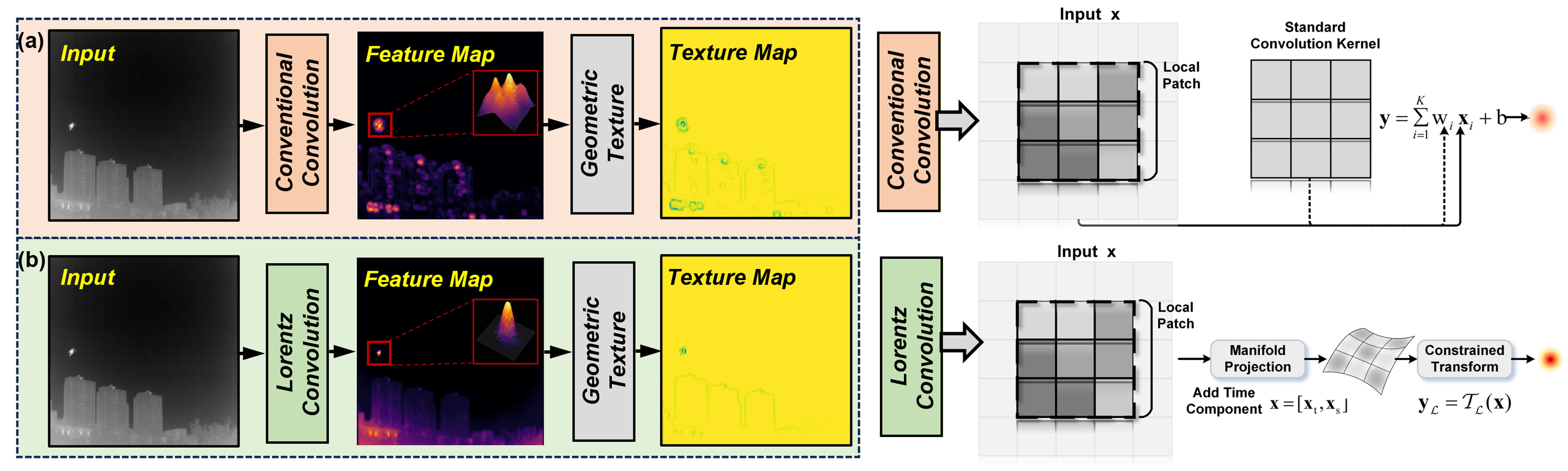}\\
  \caption{Comparison of Euclidean weighted aggregation and Lorentz manifold feature transformation. 
Conventional convolution performs local weighted summation as $y=\sum_{i=1}^{K} w_i \mathbf{x}_i+b$, 
where $\mathbf{x}_i$ and $w_i$ denote the local feature and its kernel weight, respectively, $K$ is the window size, and $b$ is the bias. 
By contrast, the Lorentz branch first maps the feature into Lorentz representation and then performs feature transformation on the Lorentz manifold, i.e., 
\(\mathbf{x}=[\mathbf{x}_t,\mathbf{x}_s]\) and \(\mathbf{y}_{\mathcal L}=\mathcal{T}_{\mathcal L}(\mathbf{x})\), 
where \(\mathbf{x}_t\) and \(\mathbf{x}_s\) denote the temporal and spatial components, respectively. Here, \(\mathcal{T}_{\mathcal L}(\cdot)\) denotes Lorentz manifold based feature transformation. 
  }\label{fig.intro}
  \end{center}
  \vspace{-0.4cm}
\end{figure*} 

To address these challenges, we propose a Lorentzian Hypergraph Network (LoHGNet) for IRSTD. The proposed network is built on the central idea of geometric encoding guided high-order relation learning. Unlike Euclidean space, Lorentz space, as a hyperbolic space with negative curvature, is more suitable for organizing hierarchical and non-uniform feature structures, thus enabling richer and more discriminative feature representations. This property is well aligned with IRSTD, where the discriminative cues of infrared small targets are usually weak and sparsely distributed, and often rely on hierarchical cues ranging from local anomalous responses to broader structural context. As shown in Fig.\ref{fig.intro}(a) and Fig.\ref{fig.intro}(b), compared with conventional convolution, Lorentz convolution produces more concentrated target responses and more clearly organized background structures. Meanwhile, the 3D response maps of the target regions further show that the features modeled by Lorentz convolution exhibit more refined non-uniform characteristics, which better align with the original target response distribution. Based on this observation, LoHGNet first performs hierarchical geometric encoding in Lorentz space and then conducts high-order contextual reasoning over hypergraphs under the guidance of the learned geometric information. Specifically, we first design a Geometric Attention Guided Lorentz Residual Convolution Module (GA-LRCM) to perform feature modeling directly on the Lorentz manifold. Based on this module, we further construct a Lorentz encoder consisting of a Lorentz Input Block (LIB), stacked GA-LRCM, and logarithmic mapping at the origin. In this way, features remain in Lorentz space throughout the encoding stage, thereby fully exploiting the hierarchical expressive capability of hyperbolic geometry. Afterward, the encoded features are mapped back to the Euclidean tangent space through logarithmic mapping, providing discriminative geometric information for subsequent relation modeling. On this basis, we further design a High-Order Relation Learning Module (HORL), which adaptively constructs hypergraphs and performs high-order relation propagation to capture the multivariate high-order contextual dependencies between targets and backgrounds. The main contributions of this work are as follows:

\begin{itemize}
\item We propose LoHGNet, which integrates hierarchical geometric representation on the Lorentz manifold with contextual relation modeling, providing a new perspective for discriminating weak targets in complex backgrounds. 
\item We design GA-LRCM, which performs progressive feature encoding under Lorentz manifold constraints to enhance the geometric representation of weak targets.
\item We design HORL, which uses geometric priors from the encoder to model high-order contextual relations between targets and complex background regions, thereby improving the discrimination between real targets and false responses in complex scenes.
\item Extensive experiments on three datasets demonstrate that LoHGNet achieves competitive performance against existing methods, validating its effectiveness and robustness in complex scenes.
\end{itemize}

\section{Related Work}
\subsection{Infrared Small Target Detection Methods}

IRSTD has long been developed along two major technical lines, namely model-driven methods and data-driven methods. Early model-driven approaches detect targets by characterizing the differences between targets and backgrounds in terms of gray-level distribution \cite{wang2012infrared}, local structure \cite{chen2013local}, and statistical properties \cite{tang2023fast}. Although such methods usually possess good physical interpretability, they are highly sensitive to background complexity, noise level, and scene variation, and are therefore prone to missed detections and false alarms in cluttered environments.

With the rapid development of deep learning, data-driven IRSTD methods have gradually become dominant. Dai et al. \cite{2021asymmetric} proposed ACM-Net, which enhances contextual interaction between local and global features through an asymmetric context modulation mechanism, thereby improving small target representation under complex backgrounds. DNA-Net \cite{li2022dense}, ALCNet \cite{dai2021attentional}, and UIU-Net \cite{wu2022uiu} further improve feature utilization efficiency from the perspectives of dense nested connections, attention modeling, and cross layer interaction, respectively. These strategies have also become commonly adopted means for enhancing feature extraction in current IRSTD research. Meanwhile, several emerging network paradigms have also been introduced into IRSTD. Mamba architectures have been explored to strengthen long-range dependency modeling while maintaining computational efficiency \cite{liu2025mou, li2026pi}. Methods built on KAN improve the discriminability between weak targets and backgrounds through learnable nonlinear mappings and dynamic convolution \cite{wu2025kpf}. In addition, frameworks incorporating large-scale foundation-model priors have also been investigated to enhance contextual understanding and fine-grained reconstruction \cite{fu2025unified}.


\subsection{Lorentz Hyperbolic Representation Learning}

Hyperbolic space, owing to its negative curvature and exponentially expanding volume, is widely regarded as well suited for representing data with intrinsic hierarchical structures. Compared with Euclidean space, it is better suited to modeling hierarchical semantics, and has therefore attracted considerable attention in representation learning. Nickel et al. \cite{nickel2018learning} showed that the Lorentz model is superior to the Poincar'e ball model in optimization efficiency and numerical stability, laying the foundation for subsequent deep learning in Lorentz space. Chen et al. \cite{chen2022fully} further extended basic neural operations to the Lorentz model and promoted the development of fully hyperbolic neural networks. More recently, hyperbolic learning has been introduced into vision tasks, and the Lorentz model has shown particular potential for deep network design due to its stable geometric operations and direct distance computation \cite{mettes2024hyperbolic}.

Although hyperbolic representation learning has shown promise in several domains, its application to IRSTD remains limited. Most existing IRSTD methods still extract features in Euclidean space and thus fail to fully exploit the advantages of hyperbolic geometry for hierarchical representation. Since IRSTD often requires jointly modeling local responses and broader structural context, mapping features into Lorentz hyperbolic space may further improve geometric structural separability.

\subsection{Hypergraph Representation Learning}
Hypergraph representation learning has become an important way to model high-order relations among multiple elements. Unlike conventional graph neural networks, which mainly describe pairwise relations through ordinary edges, hypergraphs allow one hyperedge to connect multiple nodes at the same time. This makes them more suitable for describing semantic groups and complex contextual structures in visual tasks. For this reason, hypergraph learning has gradually been introduced into deep learning and visual understanding tasks.
Feng et al. \cite{feng2019hypergraph} proposed HGNN, which introduced hypergraph convolution into deep learning and showed that hypergraphs are effective for modeling high-order correlations. As hypergraph learning has been further studied in vision tasks, later works pointed out that an image should not be treated only as a conventional graph, because this may miss high-order semantic relations among different regions \cite{han2023vision}. Hypergraph networks have since been used in visual recognition, relation modeling, and multimodal learning, largely because they support group-wise interaction in a ``node-hyperedge-node'' manner. This property makes hypergraphs a natural choice for describing the relations between targets and cluttered backgrounds.

\begin{figure*}[ht]
  \begin{center}
  \includegraphics[width= 1\linewidth]{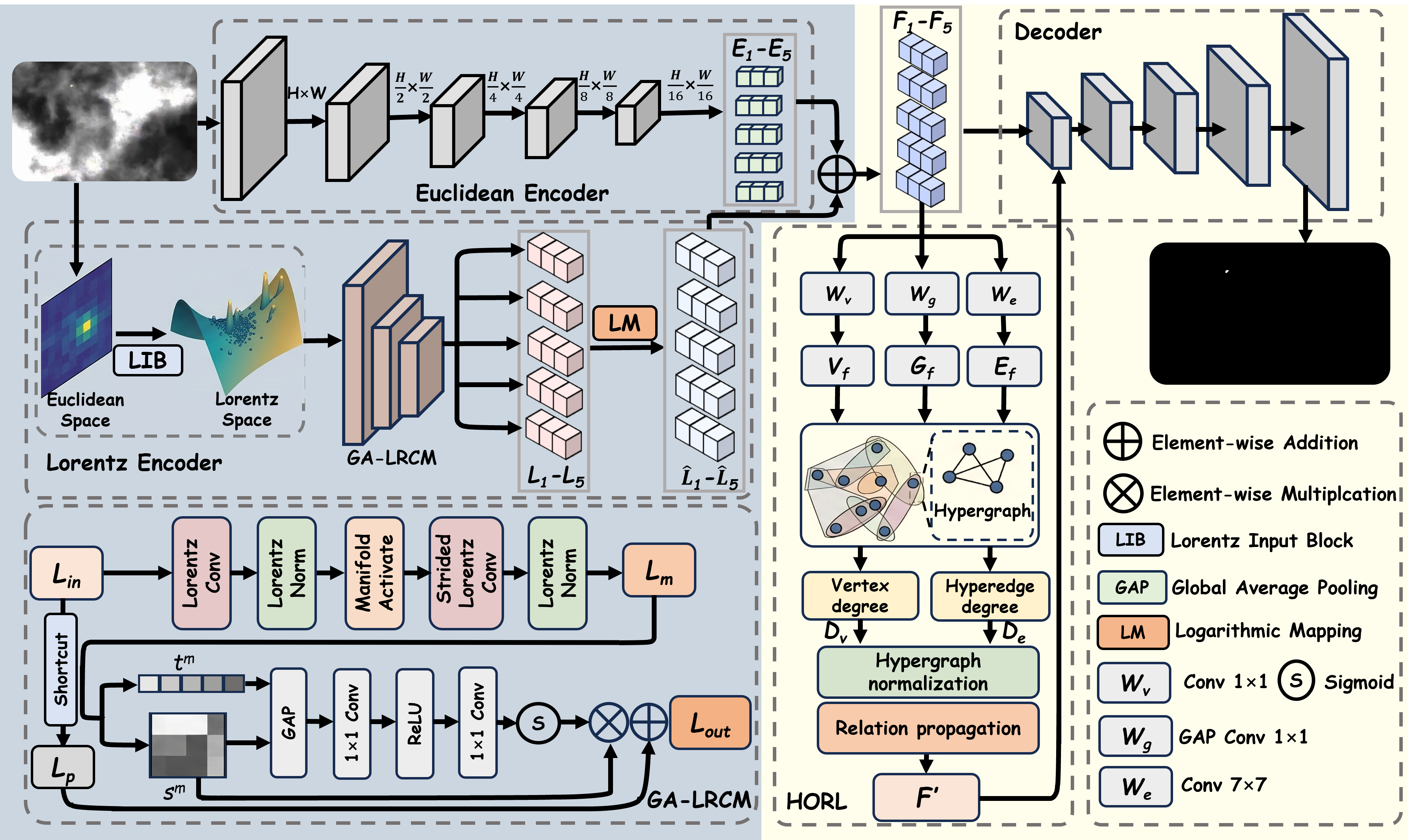}\\
  \caption{Overview of LoHGNet. The network includes a Euclidean branch for local detail preservation and a Lorentz branch for hierarchical geometric encoding through LIB and multiple GA-LRCM. HORL models high-order contextual relations between targets and complex backgrounds, followed by the decoder for final detection.
  }\label{fig.framework}
  \end{center}
\vspace{-0.6cm}
\end{figure*}

\section{Method}

\subsection{Overview}
The overall architecture of LoHGNet is shown in Fig.~\ref{fig.framework}. Given an input infrared image \(\mathbf{I} \in \mathbb{R}^{1 \times H \times W}\), the network builds a Euclidean encoding branch and a Lorentz encoding branch in parallel to extract local appearance details and hierarchical geometric information, respectively. The Euclidean branch produces multiscale features \(\{\mathbf{E}_i\}_{i=1}^{5}\) through a series of convolution and downsampling operations, where \(\mathbf{E}_i\) denotes the Euclidean feature at the \(i\)th scale.

In parallel with the Euclidean branch, the Lorentz branch represents the input feature in the Lorentz model. The Lorentz model is a common form of hyperbolic space, defined on the upper sheet of a two sheeted hyperboloid in \((n+1)\)-dimensional Minkowski space. Let a point be written as \(\mathbf{x}=[\mathbf{x}_t,\mathbf{x}_s]\), where \(\mathbf{x}_t \in \mathbb{R}\) denotes the temporal component and \(\mathbf{x}_s \in \mathbb{R}^{n}\) denotes the spatial component. Then the Lorentz manifold can be written as
\begin{equation}
\mathcal{L}_k^n=\{\mathbf{x}\in\mathbb{R}^{n+1}\mid \langle \mathbf{x},\mathbf{x}\rangle_{\mathcal L}=-k,\;\mathbf{x}_t>0\},
\end{equation}
where \(k>0\) is a constant related to the manifold curvature, and \(\langle \cdot,\cdot\rangle_{\mathcal L}\) denotes the Lorentz inner product, defined as
\begin{equation}
\langle \mathbf{x},\mathbf{y}\rangle_{\mathcal L}=-\mathbf{x}_t\mathbf{y}_t+\mathbf{x}_s^{\top}\mathbf{y}_s.
\end{equation}

In the Lorentz model, the first dimension is called the time component, and the remaining dimensions are called the space component. The time component and the space component are not independent, but jointly satisfy the manifold constraint. Under this constraint, the time component is uniquely determined by the space component, that is
\begin{equation}
\mathbf{x}_t=\sqrt{k+\|\mathbf{x}_s\|_2^2}.
\end{equation}

This property allows the Lorentz model to organize hierarchical relations and nonuniform distributions while preserving geometric constraints. For IRSTD, target discrimination often depends on weak local responses, sparse cues, and differences from the surrounding background. Based on the above definition, we use LIB to map the input feature onto the Lorentz manifold. Specifically, LIB adds a time dimension to the spatial component of the input feature and then applies manifold projection to obtain a valid representation that satisfies the Lorentz constraint, thereby mapping the feature from Euclidean space to Lorentz representation space. Let the mapped feature be denoted as \(\mathbf{L}=[\mathbf{l}_t,\mathbf{l}_s]\), where \(\mathbf{l}_t\) and \(\mathbf{l}_s\) denote the time component and the space component, respectively. This process provides the initial manifold representation for the subsequent geometric encoding. Then, multiple GA-LRCM perform progressive feature transformation and hierarchical encoding directly on the Lorentz manifold, producing multiscale hyperbolic features \(\{\mathbf{L}_i\}_{i=1}^{5}\). In this way, the features remain in Lorentz space throughout the encoding stage, which helps preserve target related cues and their geometric differences from complex backgrounds.

After Lorentz encoding, a logarithmic map at the origin is applied to the hyperbolic features at all scales, so that the manifold representations are converted into the Euclidean tangent space. Let \(\mathbf{o}\) denote the reference origin on the Lorentz manifold, and let \(\mathbf{L}\) denote a feature map defined on the same manifold. In implementation, the logarithmic map is applied pointwise to \(\mathbf{L}\), so that each Lorentz feature point is mapped into the Euclidean tangent space at \(\mathbf{o}\). For a feature point on the manifold, the geodesic distance between \(\mathbf{o}\) and \(\mathbf{L}\) is defined as
\begin{equation}
d_{\mathcal L}(\mathbf{o},\mathbf{L}) = \sqrt{k}\,\mathit{arcosh}\left(-\frac{\langle \mathbf{o},\mathbf{L}\rangle_{\mathcal L}}{k}\right),\label{eq.dl}
\end{equation}
where \(d_{\mathcal L}(\mathbf{o},\mathbf{L})\) denotes the hyperbolic geodesic distance, and \(\mathit{arcosh}(\cdot)\) is the inverse hyperbolic cosine function. The tangent direction from the origin to \(\mathbf{L}\) is written as
\begin{equation}
\mathbf{v} = \mathbf{L} + \frac{1}{k}\langle \mathbf{o},\mathbf{L}\rangle_{\mathcal L}\,\mathbf{o},\label{eq.v}
\end{equation}
where \(\mathbf{v}\) is the tangent vector at the origin. The logarithmic map is then given by
\begin{equation}
\mathit{log}_{\mathbf{o}}(\mathbf{L}) = d_{\mathcal L}(\mathbf{o},\mathbf{L})\frac{\mathbf{v}}{\|\mathbf{v}\|_2}.\label{eq.log}
\end{equation}

From Eq.~\ref{eq.dl} to Eq.~\ref{eq.log}, features on the Lorentz manifold are mapped into the Euclidean tangent space while keeping their geometric information. In practice, after logarithmic mapping, we discard the temporal component and keep only the spatial component. This component is then fused, scale by scale, with the corresponding Euclidean feature. Let \(\hat{\mathbf{L}}_i\) denote the spatial component after logarithmic mapping at the \(i\)th scale. The fused feature is written as
\begin{equation}
\mathbf{F}_i = \hat{\mathbf{L}}_i + \mathbf{E}_i.
\end{equation}
Here, \(\mathbf{F}_i\) denotes the fused feature at the \(i\)th scale. In this way, the network keeps the local intensity information from the Euclidean branch together with the geometric information from the Lorentz branch. HORL is then introduced for hypergraph construction and high-order contextual reasoning. The resulting representation, together with the shallow fused features, is finally fed into a progressive decoder to produce the final detection result.

\subsection{Geometric Attention Guided Lorentz Residual Convolution Module}
GA-LRCM serves as the core module of the Lorentz encoding branch. It is designed to support stable hierarchical feature encoding on the Lorentz manifold while preserving the hyperbolic geometric constraint. As shown in Fig.~\ref{fig.framework}, let the input of the module be the Lorentz feature \(\mathbf{L}_{\mathrm{in}}=[\mathbf{t}_{\mathrm{in}},\mathbf{s}_{\mathrm{in}}]\), where \(\mathbf{t}_{\mathrm{in}}\) and \(\mathbf{s}_{\mathrm{in}}\) denote the temporal and spatial components, respectively. In the main branch, Lorentz convolution, Lorentz normalization, and manifold activation are applied in sequence to extract local hyperbolic discriminative features. A strided Lorentz convolution is then used to reduce the spatial resolution and increase the channel dimension. To avoid numerical instability of the temporal component during convolutional updates, a lower bound is used in implementation, and a valid temporal coordinate is reconstructed from the spatial component according to the Lorentz constraint. This design ensures that the propagated feature always remains on the Lorentz manifold.

On this basis, a geometric attention mechanism is further introduced into GA-LRCM to enhance weak target responses. Let the output of the main branch before attention be \(\mathbf{L}_{m}=[\mathbf{t}^{m},\mathbf{s}^{m}]\). Following the implementation, channel statistics are extracted from the temporal and spatial components, respectively. These statistics are then used to generate channel wise modulation weights:
\begin{equation}
\boldsymbol{\alpha}=\mathit{\sigma}\!\left(\mathbf{W}_2\,\mathit{\delta}\!\left(\mathbf{W}_1\left[\mathit{GAP}(\mathbf{t}^{m}), \mathit{GAP}(\mathbf{s}^{m})\right]\right)\right),\label{eq.weight}
\end{equation}
where \(\mathit{GAP}(\cdot)\) denotes global average pooling, \(\mathit{\delta}(\cdot)\) denotes the ReLU activation, \(\mathit{\sigma}(\cdot)\) denotes the Sigmoid function, \(\mathbf{W}_1\) and \(\mathbf{W}_2\) denote two learnable \(1\times1\) convolutional transforms, and \(\boldsymbol{\alpha}\) is the geometric attention weight. Unlike conventional Euclidean attention, the weight in Eq.~\ref{eq.weight} is generated from the temporal response and spatial magnitude of Lorentz features, which makes it more suitable for emphasizing target related cues in the hierarchical representation. 

Since downsampling requires both scale and channel alignment, we use projection based residual alignment in the shortcut branch. Let the aligned shortcut output be \(\mathbf{L}_{p}=[\mathbf{t}^{p},\mathbf{s}^{p}]\). During feature fusion, only the spatial components from the attention refined main branch and the shortcut branch are combined, and the temporal component is then reconstructed according to the Lorentz constraint. The final output feature is written as
\begin{equation}
\mathbf{s}^{f} = \boldsymbol{\alpha} \odot \mathbf{s}^{m} + \mathbf{s}^{p},\quad
\mathbf{t}^{f} = \sqrt{k + \|\mathbf{s}^{f}\|_2^2},\quad
\end{equation}
where \(\odot\) denotes element wise multiplication, \(\mathbf{s}^{f}\) and \(\mathbf{t}^{f}\) denote the fused spatial and temporal components, respectively, and \(\mathbf{L}_{\mathrm{out}} = [\mathbf{t}^{f},\mathbf{s}^{f}]\) denotes the final output of GA-LRCM. This strategy, namely spatial fusion followed by temporal reconstruction, keeps the feature under the Lorentz constraint, avoids the distortion caused by directly adding full Lorentz vectors in Euclidean form, and preserves residual information at the same time. As a result, weak target responses are less likely to be smoothed out during progressive downsampling. By stacking multiple GA-LRCM blocks, the network gradually builds multiscale hyperbolic representations and provides stable geometric priors for the subsequent high-order relation modeling.

\subsection{High-Order Relation Learning}
After obtaining the deep feature, LoHGNet further employs HORL to model high-order contextual relations between target and background regions. As shown in Fig.~\ref{fig.framework}, let the input feature be \(\mathbf{F}\). Unlike conventional convolution, which aggregates information only within a local receptive field, HORL models relations over a broader spatial range. To do this, the two dimensional feature map is first rearranged into a set of node features. Then, three learnable mappings are used to extract the components needed for hypergraph construction, including vertex features, global guidance, and hyperedge features. Specifically, the input feature is processed by three learnable branches, parameterized by \(\mathbf{W}_v\), \(\mathbf{W}_g\), and \(\mathbf{W}_e\), to generate the vertex representation \(\mathbf{V}_f\), the global guidance matrix \(\mathbf{G}_f\), and the hyperedge representation \(\mathbf{E}_f\), respectively. In implementation, \(\mathbf{W}_v\) is realized by a \(1\times1\) convolution to extract compact vertex features, \(\mathbf{W}_g\) uses global average pooling followed by a \(1\times1\) convolution to provide scene level guidance, and \(\mathbf{W}_e\) is implemented by a larger kernel convolution to capture broader spatial patterns for hyperedge construction. Based on these components, the vertex hyperedge incidence matrix is constructed as
\begin{equation}
\mathbf{H}=\left|\mathbf{V}_f \mathbf{G}_f \mathbf{V}_f^{\top}\mathbf{E}_f\right|.
\end{equation}

Here, the matrix multiplication combines vertex correlation, global guidance, and hyperedge structure. Each element of \(\mathbf{H}\) represents the connection strength between a vertex and a hyperedge, and \(|\cdot|\) denotes the element wise absolute value used to ensure nonnegative weights. Since \(\mathbf{H}\) is generated directly from the input feature, HORL can build different high-order relation patterns for different scenes. This helps describe the contextual relations between targets and complex backgrounds more effectively.

In complex infrared scenes, the initial matrix often contains many weak or invalid connections caused by background edges, bright clutter, and noise. If these connections are directly used for propagation, background interference may spread through the hypergraph and reduce the discriminability of target regions. To reduce this effect, a sparsity constraint is applied before hypergraph propagation, so that only high confidence vertex hyperedge connections are kept:
\begin{equation}
\mathbf{H}_s(i,j)=\mathbf{H}(i,j)\cdot \mathbb{I}\!\left(\mathbf{H}(i,j)>\lambda\,\mathit{mean}(\mathbf{H})\right),\label{eq.hs}
\end{equation}
where \(\mathbf{H}_s\) denotes the sparsified vertex hyperedge matrix, \(\lambda\) is the sparsity factor, \(\mathit{mean}(\mathbf{H})\) denotes the average value of \(\mathbf{H}\), and \(\mathbb{I}(\cdot)\) is the indicator function, which equals 1 when the condition is satisfied and 0 otherwise. The vertex degree matrix and the hyperedge degree matrix are then computed as
\begin{equation}
\mathbf{D}_v(i,i)=\sum_j \mathbf{H}_s(i,j),\quad
\mathbf{D}_e(j,j)=\sum_i \mathbf{H}_s(i,j),
\end{equation}
where \(\mathbf{D}_v\) and \(\mathbf{D}_e\) denote the vertex degree matrix and the hyperedge degree matrix, respectively. They are used for hypergraph normalization. Based on them, the normalized hypergraph interaction matrix is written as
\begin{equation}
\mathbf{P}_H=\mathbf{D}_v^{-1/2}\mathbf{H}_s\mathbf{D}_e^{-1}\mathbf{H}_s^{\top}\mathbf{D}_v^{-1/2}.
\end{equation}

Here, the superscript \(-1/2\) denotes inverse square root normalization on the vertex degree matrix, so that vertices with many connections do not contribute too much during propagation. The superscript \(-1\) denotes inverse normalization on the hyperedge degree matrix, so that hyperedges covering many vertices do not produce overly strong influence. Therefore, \(\mathbf{P}_H\) describes the normalized high order interactions on the hypergraph. Based on \(\mathbf{P}_H\), relation propagation is performed as
\begin{equation}
\mathbf{F}'=\mathbf{F}\mathbf{\Theta}-\mathbf{P}_H\mathbf{F}\mathbf{\Theta}, \label{eq12}
\end{equation}
where \(\mathbf{F}\) denotes the input feature, \(\mathbf{\Theta}\) is a learnable linear transform, and \(\mathbf{F}'\) denotes the output feature after high-order relational propagation. 

\section{Experiments}
\subsection{Experimental Setup}

1) \textbf{\textit{Datasets}}: Experiments are conducted on three widely used IRSTD datasets, including NUDT-SIRST~\cite{li2022dense}, NUAA-SIRST~\cite{2021asymmetric}, and IRSTD-1K~\cite{zhang2022isnet}. For fair comparison, the official training and testing splits released in the original papers are strictly followed for all datasets.

2) \textbf{\textit{Evaluation Metrics}}: We adopt several standard metrics for evaluation, including Intersection over Union (\(\mathrm{IoU}\)), normalized IoU (\(\mathrm{nIoU}\)), F-measure (\(F\)), Probability of Detection (\(P_d\)), and False Alarm Rate (\(F_a\)). Among them, \(\mathrm{IoU}\) measures the pixel level overlap between the prediction and the ground truth, while \(\mathrm{nIoU}\) reduces the influence of target scale variation and provides a more balanced evaluation across different target sizes. \(F\) evaluates miss detection and false alarms at the pixel level. \(P_d\) measures the proportion of correctly detected targets, and \(F_a\) measures the proportion of falsely detected pixels in the whole image, reflecting the model's ability to suppress background interference.

3) \textbf{\textit{Implementation Details}}: The model is trained for 1000 epochs with a batch size of 4. All input images are first normalized and then cropped into patches of size $256 \times 256$. The implementation is based on PyTorch, and all experiments are carried out on an NVIDIA RTX 4090D GPU.

\begin{table*}[t]
\centering
\caption{Overall ablation study on NUDT-SIRST, NUAA-SIRST, and IRSTD-1K.}
\label{tab,overall_ablation}
\setlength{\tabcolsep}{6pt}
\renewcommand{\arraystretch}{1.15}
\begin{tabular}{c ccc ccc ccc ccc}
\toprule
\multirow{2}{*}{Config} & \multicolumn{3}{c}{Module} & \multicolumn{3}{c}{NUDT-SIRST} & \multicolumn{3}{c}{NUAA-SIRST} & \multicolumn{3}{c}{IRSTD-1K} \\
\cmidrule(lr){2-4} \cmidrule(lr){5-7} \cmidrule(lr){8-10} \cmidrule(lr){11-13}
& Conv Branch & GA-LRCM & HORL & IoU & nIoU & F & IoU & nIoU & F & IoU & nIoU & F \\
\midrule
1 & $\times$      & $\times$      & $\times$      & 91.04\%          & 91.12\%          & 94.07\%          & 73.95\%          & 76.32\%          & 84.15\%          & 63.21\%          & 60.48\%          & 76.49\%          \\
2 & $\checkmark$  & $\times$      & $\times$      & 92.85\%          & 92.59\%          & 95.27\%          & 75.20\%          & 77.59\%          & 85.62\%          & 64.96\%          & 62.20\%          & 77.21\%          \\
3 & $\times$      & $\checkmark$  & $\times$      & 93.23\%          & 93.42\%          & 96.31\%          & 75.71\%          & 77.87\%          & 85.94\%          & 65.47\%          & 62.61\%          & 77.42\%          \\
4 & $\times$      & $\times$      & $\checkmark$  & 92.74\%          & 92.81\%          & 95.36\%          & 75.13\%          & 77.45\%          & 85.48\%          & 65.03\%          & 61.95\%          & 77.26\%          \\
5 & $\checkmark$  & $\checkmark$  & $\times$      & 93.99\%          & 94.06\%          & 97.15\%          & 75.97\%          & 78.84\%          & 86.73\%          & 66.48\%          & 63.04\%          & 77.81\%          \\
6 & $\times$  & $\checkmark$      & $\checkmark$  & 94.58\%          & 94.80\%          & 97.19\%          & 76.20\%          & 79.22\%          & 86.86\%          & 65.99\%          & 64.12\%          & 78.08\%          \\
\rowcolor{gray!15}
7 & $\checkmark$  & $\checkmark$  & $\checkmark$  & \textbf{95.61\%} & \textbf{95.50\%} & \textbf{97.67\%} & \textbf{77.26\%} & \textbf{79.91\%} & \textbf{86.80\%} & \textbf{68.03\%} & \textbf{64.26\%} & \textbf{80.28\%} \\
\bottomrule
\end{tabular}\label{table_zongxiaorong}
\end{table*}
\begin{figure*}[h]
  \begin{center}
  \includegraphics[width= 0.9\linewidth]{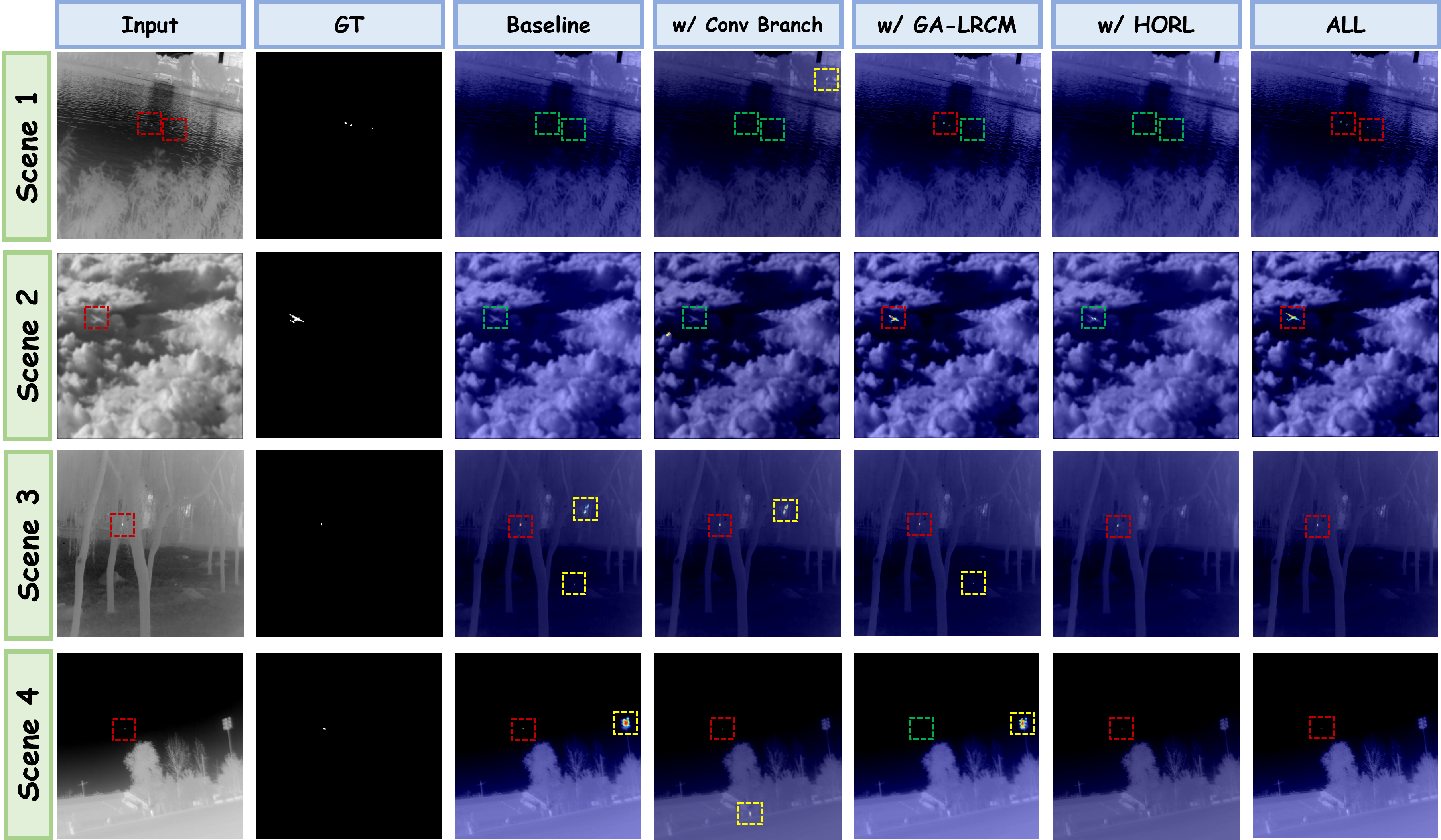}\\
  \caption{Visualization comparison of feature maps for different ablation settings, where red boxes denote correctly detected targets, green boxes denote missed targets, and yellow boxes denote false alarms.
  }\label{fig.xiaorongduibi}
  \end{center}
\vspace{-0.6cm}
\end{figure*}

\subsection{Ablation Study}

\textit{1) \textbf{Ablation Study on Module Combinations}}: To comprehensively analyze the contribution of each component to the overall performance, we first conduct an ablation study on the overall architecture of LoHGNet, and the corresponding results are reported in Table~\ref{table_zongxiaorong}. Here, Conv Branch represents the introduction of the Euclidean encoder, GA-LRCM represents the introduction of the Lorentz encoder, and HORL represents the high-order relation learning module. It can be observed that the model consistently improves over the baseline when Conv Branch, GA-LRCM, or HORL is introduced individually, indicating that Euclidean detail encoding, Lorentz geometric encoding, and high-order relation modeling all contribute positively to IRSTD. In the module combination experiments, further gains are obtained when GA-LRCM is combined with Conv Branch, or when GA-LRCM and HORL are jointly introduced, showing that local detail preservation, geometric representation, and contextual relation modeling are complementary. Finally, the complete model achieves the best results on all three datasets, improving the IoU by approximately 4.57, 3.31, and 4.82 percentage points over the baseline, respectively, while also maintaining clear advantages in terms of nIoU and F measure. These results show that the proposed modules do not work independently, but instead cooperate within the overall framework. In particular, GA-LRCM improves the geometric discriminability of features, Conv Branch provides local detail information, and HORL further strengthens relation modeling under complex backgrounds, leading to stable performance gains.

Fig.~\ref{fig.xiaorongduibi} shows the visual results of the overall ablation study on four scenes. Compared with the baseline and incomplete variants, the full model gives more stable target responses and fewer false alarms across different scenes. In Scene 1, the targets are small and the background is cluttered. Only the full model detects the true targets accurately, while the other variants suffer from missed detections or false alarms. This suggests that a single module is not enough to enhance weak targets and suppress background interference at the same time. In Scene 2, the targets have relatively clear shapes but are still embedded in a complex background. The full model and the variant with GA-LRCM alone give better target responses than the other variants, showing the advantage of Lorentz geometric encoding in preserving target related cues. In Scene 3 and Scene 4, several high-response background regions are easily confused with true targets. The full model and the variant with HORL alone suppress these false responses more effectively, while the variants without HORL are more likely to produce false alarms. This shows that HORL helps distinguish real targets from background induced false responses in complex scenes.

\begin{figure*}[h]
  \begin{center}
  \includegraphics[width= 0.9\linewidth]{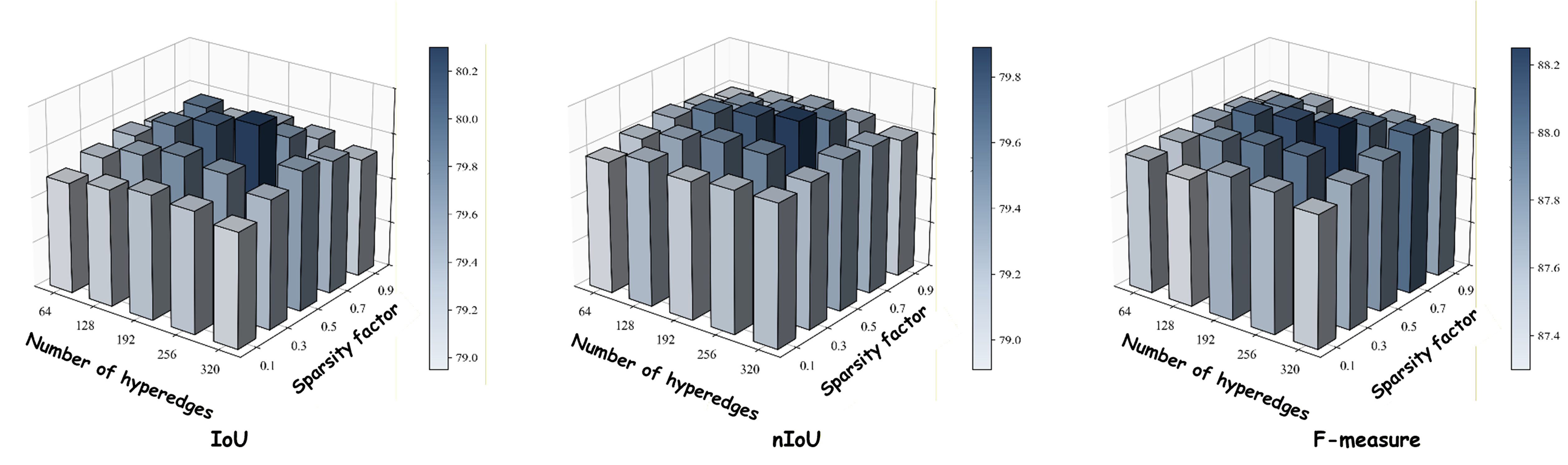}\\
  \caption{Parameter analysis of HORL in terms of mean IoU, nIoU, and F-measure under different combinations of the sparsity factor and the number of hyperedges.
  }\label{fig.canshu}
  \end{center}
\vspace{-0.6cm}
\end{figure*}

\begin{table}[t]
\centering
\scriptsize
\caption{Ablation studies of GA-LRCM and HORL on NUDT-SIRST, NUAA-SIRST, and IRSTD-1K.}
\label{tab:ablation_all}
\setlength{\tabcolsep}{3pt}
\renewcommand{\arraystretch}{1.08}
\resizebox{\columnwidth}{!}{
\begin{tabular}{l ccc ccc ccc}
\toprule
\multirow{2}{*}{Config} & \multicolumn{3}{c}{NUDT-SIRST} & \multicolumn{3}{c}{NUAA-SIRST} & \multicolumn{3}{c}{IRSTD-1K} \\
\cmidrule(lr){2-4} \cmidrule(lr){5-7} \cmidrule(lr){8-10}
& IoU & nIoU & F & IoU & nIoU & F & IoU & nIoU & F \\
\midrule
Baseline & 91.04 & 91.12 & 94.07 & 73.95 & 76.32 & 84.15 & 63.21 & 60.48 & 76.49 \\
GA-LRC w/o Residual Branch & 91.87 & 92.10 & 94.92 & 74.68 & 76.97 & 84.99 & 64.39 & 61.82 & 76.99 \\
GA-LRC w/o Geometry Attention & 92.47 & 92.35 & 95.10 & 74.93 & 77.16 & 84.83 & 64.44 & 61.96 & 77.05 \\
\rowcolor{gray!15}
GA-LRCM & \textbf{93.23} & \textbf{93.42} & \textbf{96.31} & \textbf{75.71} & \textbf{77.87} & \textbf{85.94} & \textbf{65.47} & \textbf{62.61} & \textbf{77.42} \\
HORL w/o Hypergraph & 91.64 & 91.97 & 95.04 & 74.71 & 76.40 & 84.92 & 64.69 & 60.98 & 76.90 \\
\rowcolor{gray!15}
HORL & \textbf{92.74} & \textbf{92.81} & \textbf{95.36} & \textbf{75.13} & \textbf{77.45} & \textbf{85.48} & \textbf{65.03} & \textbf{61.95} & \textbf{77.26} \\
\bottomrule
\end{tabular}}
\end{table}

\begin{figure}[h]  

  \begin{center}
  \includegraphics[width= 1.0\linewidth]{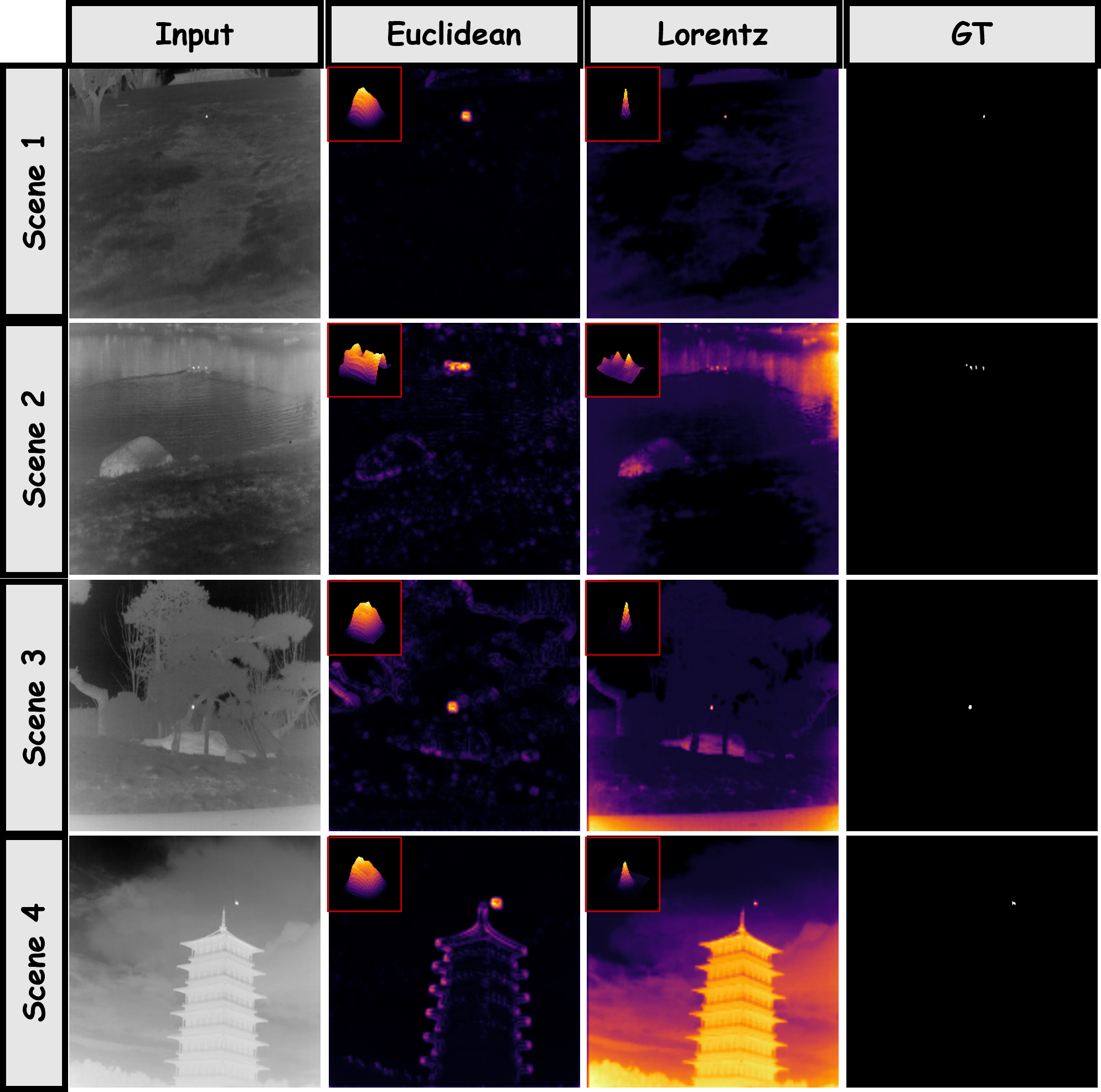}\\
  \caption{Comparison of same-scale feature maps in Euclidean space and Lorentz space.
  }\label{fig.2space_feature_map}
  \end{center}
  \vspace{-0.4cm}
\end{figure} 

\textit{2) \textbf{Ablation Study on GA-LRCM and HORL}}: To further evaluate the roles of GA-LRCM and HORL, we report the corresponding ablation results in Table~\ref{tab:ablation_all}. For GA-LRCM, removing either the residual branch or the geometry attention leads to consistent drops in IoU, nIoU, and F-measure on all three datasets, compared with the complete GA-LRCM. This shows that both components are useful for stable feature encoding, and that the full GA-LRCM gives the strongest geometric representation for weak target detection. For HORL, replacing the full module with HORL w/o Hypergraph also reduces the performance on all three datasets. This indicates that hypergraph based relation modeling is more effective than the simplified variant in capturing contextual relations between targets and complex backgrounds.

To further analyze the effect of key hyperparameters in HORL, Fig.~\ref{fig.canshu} shows the changes in the mean IoU, nIoU, and F-measure under different combinations of the sparsity factor and the number of hyperedges. The performance first improves and then becomes stable as the parameter setting changes, and better results are obtained with a moderate sparsity factor and a proper number of hyperedges. This suggests that weak sparsity is not enough to suppress invalid connections, while overly strong sparsity or too many hyperedges may introduce unnecessary relations. The best results are obtained when the sparsity factor is 0.5 and the number of hyperedges is 256.

\textit{3) \textbf{Feature Representation in Euclidean Space and Lorentz Space}}: To compare feature representations in Euclidean space and Lorentz space more directly, Fig.~\ref{fig.2space_feature_map} shows the visual results in four scenes, including the input image, the same scale feature maps in Euclidean space, the same scale feature maps in Lorentz space, the corresponding GT, and the local 3D response maps of the target regions. It can be seen that, although the feature maps in Euclidean space can highlight parts of the targets, their responses are relatively scattered in complex scenes. By contrast, the feature maps in Lorentz space show more concentrated responses around the target regions. In weak point target scenes and scenes with strong background interference, Euclidean space tends to mix nearby responses together, while Lorentz space keeps them more separated. The local 3D response maps further show that the peaks in Euclidean space are usually flatter and more spread out, whereas the peaks in Lorentz space are more compact. These results suggest that Lorentz space provides a different feature representation from Euclidean space and offers useful geometric cues for the whole network, which is helpful for distinguishing weak targets from complex backgrounds.

\begin{table*}[t]
\centering
\scriptsize
\renewcommand{\arraystretch}{1.15}
\setlength{\tabcolsep}{4.5pt}
\caption{Quantitative comparison with various methods on the NUDT-SIRST, NUAA-SIRST, and IRSTD-1K datasets in terms of IoU $(\%)$, nIoU $(\%)$, F-measure $(\%)$, $P_d$ $(\%)$, and $F_a$ $(\times 10^{-6})$. $\uparrow$/$\downarrow$ indicates that a larger/smaller value is better. Bold values denote the best result under each metric.}
\label{label.duibi_zong}
\resizebox{\linewidth}{!}{
\begin{tabular}{lccccc ccccc ccccc}
\toprule
\multirow{2}{*}{Method} 
& \multicolumn{5}{c}{NUDT-SIRST} 
& \multicolumn{5}{c}{NUAA-SIRST} 
& \multicolumn{5}{c}{IRSTD-1K} \\
\cmidrule(lr){2-6} \cmidrule(lr){7-11} \cmidrule(lr){12-16}
& IoU $\uparrow$ & nIoU $\uparrow$ & F-measure $\uparrow$ & $P_d$ $\uparrow$ & $F_a$ $\downarrow$
& IoU $\uparrow$ & nIoU $\uparrow$ & F-measure $\uparrow$ & $P_d$ $\uparrow$ & $F_a$ $\downarrow$
& IoU $\uparrow$ & nIoU $\uparrow$ & F-measure $\uparrow$ & $P_d$ $\uparrow$ & $F_a$ $\downarrow$ \\
\midrule
Top-Hat \cite{zeng2006design}        
& 20.72 & 28.98 & 33.52 & 78.41 & 166.7 
& 7.143 & 18.27 & 14.63 & 79.84 & 1012 
& 10.06 & 7.438 & 16.02 & 75.11 & 1432 \\

Max-Median \cite{deshpande1999max}   
& 4.197 & 3.674 & 7.635 & 58.41 & 36.89 
& 4.172 & 12.31 & 10.67 & 69.20 & 55.33 
& 6.998 & 3.051 & 8.152 & 65.21 & 59.73 \\

WSLCM \cite{han2020infrared}         
& 2.283 & 3.865 & 5.987 & 56.82 & 1309 
& 1.158 & 6.835 & 4.812 & 77.95 & 5446 
& 3.452 & 0.678 & 2.125 & 72.44 & 6619 \\

TTLCM \cite{han2019local}            
& 2.176 & 4.315 & 7.225 & 62.01 & 1608 
& 1.029 & 4.099 & 4.995 & 79.09 & 5899 
& 3.311 & 0.784 & 2.186 & 77.39 & 6738 \\

IPI \cite{gao2013infrared}           
& 17.76 & 15.42 & 26.94 & 74.49 & 41.23 
& 25.67 & 50.17 & 43.65 & 84.63 & 16.67 
& 27.92 & 20.46 & 35.68 & 81.37 & 16.18 \\

PSTNN \cite{zhang2019infrared}       
& 14.85 & 23.57 & 35.63 & 66.13 & 44.17 
& 30.30 & 33.67 & 39.16 & 72.80 & 48.99 
& 24.57 & 17.93 & 37.18 & 71.99 & 35.26 \\

MSLSTIPT \cite{sun2020infrared}      
& 8.342 & 10.06 & 18.26 & 47.40 & 888.1 
& 10.30 & 15.93 & 18.83 & 82.13 & 1131 
& 11.43 & 5.932 & 12.23 & 79.03 & 1524 \\
\midrule
ACM \cite{2021asymmetric}            
& 69.00 & 71.61 & 81.58 & 97.25 & 15.97 
& 68.18 & 68.83 & 81.20 & 93.54 & 32.65 
& 55.09 & 51.84 & 70.83 & 87.54 & 105.7 \\

ALC-Net \cite{dai2021attentional}       
& 71.08 & 73.83 & 83.07 & 97.14 & 11.74 
& 70.37 & 71.39 & 82.50 & 91.63 & 27.30 
& 52.55 & 51.61 & 68.75 & 87.21 & 73.96 \\

ISNet \cite{zhang2022isnet}          
& 72.06 & 73.75 & 83.76 & 94.18 & 30.54 
& 74.36 & 73.58 & 85.30 & 95.44 & 34.83 
& 65.59 & 61.74 & 79.19 & \textbf{93.60} & 45.97 \\

DNA-Net \cite{li2022dense}           
& 84.81 & 85.59 & 91.77 & 98.52 & 5.814 
& 75.42 & 72.58 & 85.96 & 96.20 & \textbf{12.34} 
& 62.76 & 58.57 & 77.13 & 91.16 & 17.61 \\

UIU-Net \cite{wu2022uiu}             
& 93.48 & 93.89 & 96.63 & 98.31 & 7.79 
& 75.74 & 77.47 & 86.20 & 96.19 & 29.42 
& 65.37 & 64.13 & 78.62 & 91.92 & 16.85 \\

RDIAN \cite{sun2023receptive}        
& 83.06 & 83.28 & 90.75 & 97.78 & 14.59 
& 71.44 & 75.05 & 83.34 & 94.30 & 35.74 
& 63.30 & 63.35 & 77.52 & 92.59 & 38.43 \\

MTU-Net \cite{wu2023mtu}             
& 71.18 & 70.39 & 84.00 & 95.77 & 25.30 
& 69.73 & 68.20 & 83.18 & 95.44 & 25.57 
& 63.22 & 59.66 & 77.51 & 88.21 & \textbf{6.083} \\

AGPCNet \cite{zhang2023attention}    
& 90.22 & 91.65 & 94.86 & 99.15 & 5.171 
& 74.06 & 75.77 & 85.09 & 95.06 & 19.14 
& 65.55 & 62.42 & 79.18 & 92.59 & 44.60 \\

SeRankDet \cite{dai2024pick}         
& 85.21 & 88.27 & 92.16 & 98.36 & 3.544 
& 76.07 & 75.53 & 86.44 & 96.20 & 19.61 
& 65.38 & 57.53 & 79.04 & 77.82 & 15.97 \\

SCTransNet \cite{yuan2024sctransnet} 
& 93.08 & 93.79 & 96.32 & 98.94 & 7.170 
& 75.72 & 77.33 & 86.05 & 96.58 & 21.54 
& 63.46 & 62.07 & 77.09 & 91.58 & 18.28 \\

GCLNet \cite{shen2025graph}          
& 87.71 & 87.47 & 93.45 & 97.67 & 5.791 
& 76.34 & 78.81 & 86.61 & 95.23 & 22.49 
& 65.35 & 63.41 & 79.04 & 92.59 & 28.70 \\

HDNet \cite{xu2025hdnet}             
& 83.81 & 84.93 & 91.19 & 98.10 & 9.675 
& 75.65 & 74.22 & 85.97 & 96.54 & 24.52 
& 65.57 & 60.35 & 79.20 & 91.84 & 11.99 \\

HaarTransNet \cite{fan2026haartransnet}             
& 93.85 & 94.34 & 96.72 & 99.05 & 2.942 
& 75.42 & 77.26 & 86.21 & 92.86 & 12.50 
& 66.65 & 64.04 & 80.03 & 92.62 & 34.86 \\

\rowcolor{gray!15}
LoHGNet                           
& \textbf{95.61} & \textbf{95.50} & \textbf{97.67} & \textbf{99.26} & \textbf{1.609} 
& \textbf{77.26} & \textbf{79.92} & \textbf{86.80} & \textbf{96.58} & 16.33 
& \textbf{68.03} & \textbf{64.26} & \textbf{80.28} & 92.59 & 25.75 \\
\bottomrule
\end{tabular}}
\end{table*}

\begin{figure*}[h]
  \begin{center}
  \includegraphics[width= 0.9\linewidth]{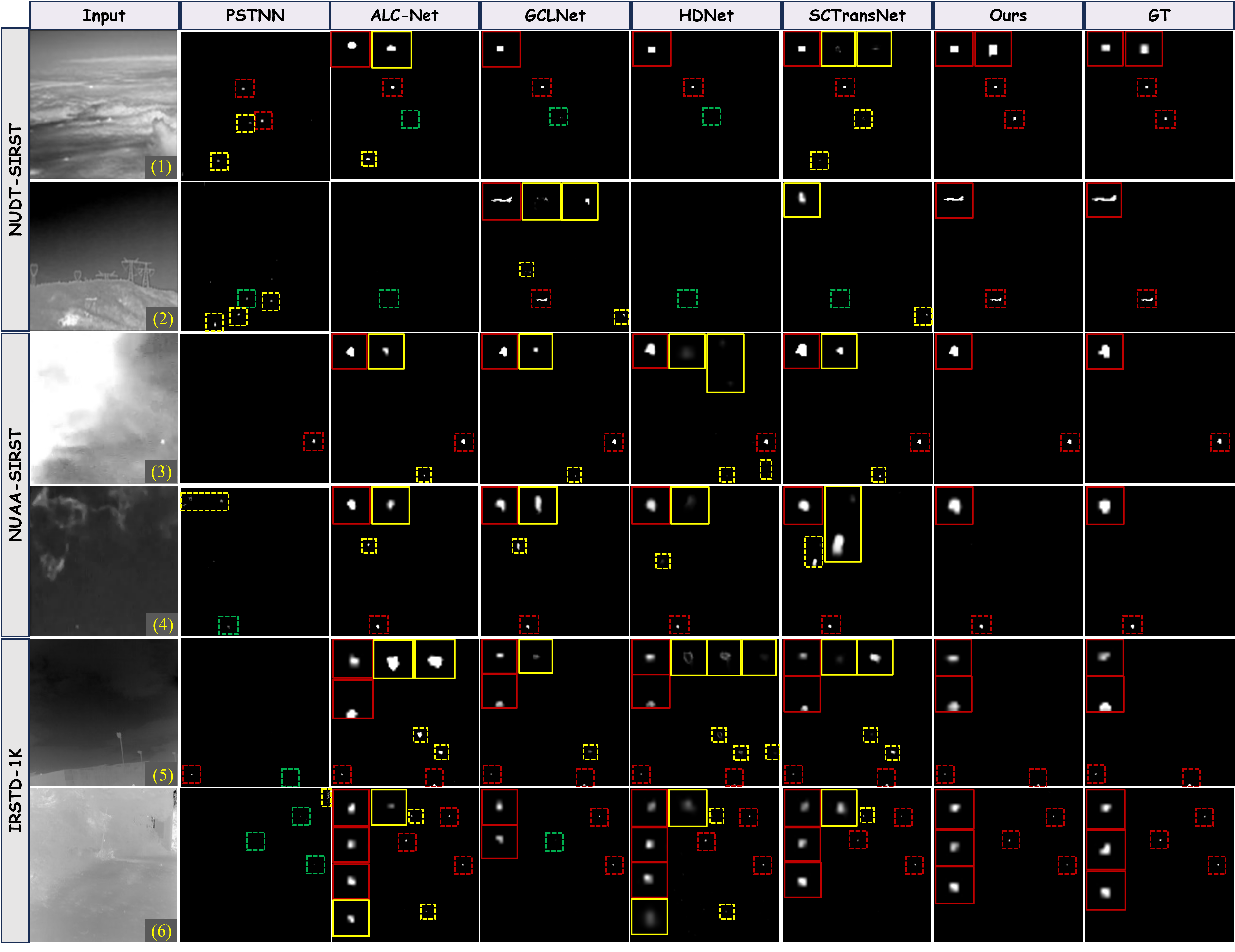}\\
  \caption{Qualitative comparison with competing methods on the NUDT-SIRST, NUAA-SIRST, and IRSTD-1K datasets.}
  \label{fig.Zongduibi}
  \end{center}
\vspace{-0.6cm}
\end{figure*}

\subsection{Comparison with State-of-the-Arts}

To validate the effectiveness of the proposed method, we conducted a comparison between LoHGNet and a series of representative IRSTD methods on three datasets. The compared methods include seven traditional algorithms, namely Top-Hat \cite{zeng2006design}, Max-Median \cite{deshpande1999max}, WSLCM \cite{han2020infrared}, TTLCM \cite{han2019local}, IPI \cite{gao2013infrared}, PSTNN \cite{zhang2019infrared} , and MSLSTIPT \cite{sun2020infrared}. In addition, thirteen deep learning-based methods were selected for comparison, including ACM \cite{2021asymmetric}, ALC-Net \cite{dai2021attentional}, ISNet \cite{zhang2022isnet}, DNA-Net \cite{li2022dense}, UIU-Net \cite{wu2022uiu}, RDIAN \cite{sun2023receptive}, MTU-Net \cite{wu2023mtu}, AGPCNet \cite{zhang2023attention}, SeRankDet \cite{dai2024pick}, SCTransNet \cite{yuan2024sctransnet}, GCLNet \cite{shen2025graph}, HDNet \cite{xu2025hdnet}, and HaarTransNet \cite{fan2026haartransnet}. Furthermore, comprehensive comparisons from both quantitative and qualitative perspectives were carried out to demonstrate the superior performance of LoHGNet on the IRSTD task.

\textbf{Quantitative Analysis:} Table \ref{label.duibi_zong} presents the quantitative comparison results between LoHGNet and various representative methods. It can be observed that deep learning-based methods generally outperform traditional methods. We attribute this to the fact that traditional methods usually rely on manually designed local contrast cues or prior assumptions, and their fixed parameter settings often lead to insufficient robustness under complex backgrounds and dim small targets. In contrast, deep learning-based methods are able to learn more discriminative multi-scale representations through end-to-end training, and therefore usually achieve better performance in terms of detection accuracy. Compared with existing deep learning-based methods, LoHGNet achieves competitive results on all three datasets. On NUDT-SIRST, LoHGNet obtains the best results in all five metrics. On the NUAA-SIRST and IRSTD-1K datasets, LoHGNet also achieves the highest IoU, nIoU, and F-measure among all competing methods. For the relatively more fluctuating metrics, namely $P_d$ and $F_a$, taking ISNet, which achieves the best $P_d$ on IRSTD-1K, as an example, although LoHGNet yields a 1.01\% lower $P_d$, it achieves a reduction of $20.22 \times 10^{-6}$ in $F_a$, indicating that our method provides a better balance between detection capability and false-alarm suppression.

\textbf{Qualitative Analysis:} To further visually validate the effectiveness of the proposed method, Fig.~\ref{fig.Zongduibi} presents the qualitative comparison results of LoHGNet against several competing methods. In the figure, red boxes denote correct detections, yellow boxes denote false alarms, and green boxes denote missed detections. From the overall results, it can be observed that traditional methods generally suffer from high missed-detection and false-alarm rates in complex scenes, which is consistent with the quantitative results in Table \ref{label.duibi_zong}. In contrast, deep learning-based methods exhibit better overall performance, but still show varying limitations in preserving dim small targets, recovering target structures, and suppressing false alarms under complex backgrounds.

For dim point-like target scenarios, such as Fig.~\ref{fig.Zongduibi}(1) and (3), several methods produce false alarms, and missed detections. By contrast, LoHGNet is able to localize the real targets more accurately, indicating its stronger discriminative capability for dim small targets. We attribute this mainly to the hierarchical geometric modeling performed by GA-LRCM in the Lorentz space, which more effectively enlarges the representation gap between targets and backgrounds, thereby preventing weak targets from being overwhelmed by complex backgrounds during feature propagation. For targets with certain structural characteristics, such as Fig.~\ref{fig.Zongduibi}(2), some competing methods can detect the target, but they still produce incomplete target structures and additional false alarms, whereas the predictions of LoHGNet are more compact and complete, and are closer to the GT. This suggests that LoHGNet is capable of not only detecting the target, but also more accurately recovering its spatial structure. 

For complex backgrounds and strong interference, such as those in Fig.~\ref{fig.Zongduibi}(5) and (6), the background contains bright scattering regions and several pseudo target responses, causing most methods to produce false alarms to varying degrees. In contrast, LoHGNet still maintains stable responses around the true targets while suppressing unrelated activations more effectively, showing better background suppression. We attribute this advantage mainly to HORL, which captures high-order relations among deep features and helps the model better distinguish true targets from background induced pseudo responses.

\section{Conclusion}
In this paper, we propose LoHGNet, an IRSTD network that combines Lorentz geometric encoding with high-order relation learning. In the encoder, GA-LRCM operates on features in Lorentz space to enhance geometric representation and support stable feature propagation, thereby improving weak target representation. HORL is then applied in the Euclidean tangent space, where hypergraph construction and high-order relation propagation are used to model contextual relations between targets and background regions. With this design, LoHGNet introduces a feature learning scheme that is different from conventional IRSTD methods.

Experimental results on three datasets show that the proposed method achieves competitive performance and gives more stable results in scenes with strong background interference. The ablation studies and visualization results further support the roles of the key modules. In particular, GA-LRCM improves the geometric representation of weak targets, while HORL helps suppress false responses and model target background relations more effectively. In future work, we will further study how Lorentz space can be combined with other network architectures for IRSTD.

\bibliographystyle{IEEEtran}
\bibliography{main}

\vfill

\end{document}